\pdfoutput=1
\documentclass[10pt, a4paper]{article}

\usepackage{lrec-coling2024} 

\usepackage{booktabs}
\usepackage{multirow}

\title{Silver Retriever: Advancing Neural Passage Retrieval\\for Polish Question Answering}

\name{
Piotr Rybak, Maciej Ogrodniczuk
}

\address{
Institute of Computer Science, Polish Academy of Sciences\\
ul. Jana Kazimierza 5, 01-248 Warsaw, Poland\\
\texttt{\{firstname.lastname\}@ipipan.waw.pl}
}

\abstract{
Modern open-domain question answering systems often rely on accurate and efficient retrieval components to find passages containing the facts necessary to answer the question. Recently, neural retrievers have gained popularity over lexical alternatives due to their superior performance. However, most of the work concerns popular languages such as English or Chinese. For others, such as Polish, few models are available. In this work, we present Silver Retriever, a neural retriever for Polish trained on a diverse collection of manually or weakly labeled datasets. Silver Retriever achieves much better results than other Polish models and is competitive with larger multilingual models. Together with the model, we open-source five new passage retrieval datasets.
\\ \newline 
}

\begin{document}

\maketitleabstract

\section{Introduction}
\label{sec:intro}

In modern open-domain question-answering systems, especially those based on Large Language Models (LLMs), it is crucial to effectively and accurately find relevant passages to correctly answer user questions and avoid hallucinations. Traditional lexical approaches such as BM25  \cite{Robertson2009ThePR} have long served as the basis for retrieval systems. However, these methods have limitations, such as being restricted to returning passages containing the exact query terms and ignoring word order.

Recently, the neural retrievers based on pre-trained language models and fine-tuned on large datasets of question-passage pairs, have significantly outperformed their lexical counterparts \citep{karpukhin-etal-2020-dense}. While progress has been focused on languages such as English and Chinese, the less popular languages such as Polish remain unexplored.

In response to this gap, we present the Silver Retriever,\footnote{We publish the model at: \url{https://hf.co/ipipan/silver-retriever-base-v1}} a Polish neural retrieval model trained on a diverse collection of both manually and weakly labeled datasets. We evaluate the model on three retrieval tasks and demonstrate that the Silver Retriever shows significant performance improvements over existing Polish models, and competes favorably with larger multilingual models.

We also open-source five new passage retrieval datasets,\footnote{We publish the datasets at: \url{https://hf.co/datasets/ipipan/maupqa}} consisting of over 500,000 questions, which will enable further research and development in the field of Polish question answering.

\section{Related Work}
\label{sec:related}

There are almost no models for Polish passage retrieval. The only available Polish neural retriever is HerBERT-QA \citep{rybak-2023-maupqa}, published together with the MAUPQA dataset on which it was trained. \citet{wojtasik2023beirpl} and \citet{polqa} have trained neural retrievers on other Polish datasets, but haven't released them.

\citet{9945218} trained a family of sentence transformers (ST-DistilRoBERTa and ST-MPNet) on an automatically constructed dataset of paraphrase pairs from sentence-aligned bilingual text corpora. While these models were trained for sentence similarity, they can also be used for retrieval.

Apart from Polish resources, there are several multilingual models. LaBSE \citep{feng-etal-2022-language} is trained to produce similar representations for bilingual sentence pairs that are translations of each other. mContriever \citep{izacard2021contriever} is a multilingual dense retriever trained with an unsupervised contrastive learning objective and fine-tuned on the MS MARCO dataset \citep{nguyen2016ms}. The E5 model \citep{wang2022text} is also trained with a contrastive objective, but on a weakly-labeled large-scale selection of text pairs and later fine-tuned on labeled datasets. SentenceTransformer \citep{reimers-2019-sentence-bert} is a library for training sentence transformers. While they focus mainly on English models, they also include several multilingual models, with MiniLM-L12-v2 being the most popular.\footnote{\url{https://hf.co/sentence-transformers/paraphrase-multilingual-MiniLM-L12-v2}}

\begin{table*}[!ht]
\renewcommand*{\arraystretch}{1.05}
\setlength{\tabcolsep}{8.5pt}
\centering
\begin{tabular}{l|rrr|rrr}
    \toprule
    \multirow{2}{*}{\bf{Dataset}} & \multicolumn{3}{c|}{\bf{\# Questions}} & \multicolumn{3}{c}{\bf{\# Passages}} \\
    & \bf{Positive} & \bf{Negative} & \bf{Total} & \bf{Positive} & \bf{Negative} & \bf{Total} \\
    \midrule
    \multicolumn{7}{c}{\bf{Existing datasets}} \\
    \midrule
    PolQA                &      4,591 &      4,961 &      5,000 &     27,131 &     34,904 &     62,035 \\
    MAUPQA               &    385,827 &    366,506 &    385,895 &    395,416 &  2,429,602 &  2,824,978 \\
    PoQuAD              &     46,161 &     51,967 &     56,588 &     46,187 &    299,865 &    346,052 \\
    PolevalPairs2021    &      1,347 &      1,913 &      1,977 &      2,088 &     17,608 &     19,690 \\
    \midrule
    \multicolumn{7}{c}{\bf{New datasets}} \\
    \midrule
    1z10                &     18,593 &     21,353 &     22,835 &     22,014 &    139,471 &    160,850 \\
    GPT3.5-CC        &     29,591 &     29,484 &     29,591 &     29,720 &    251,959 &    281,679 \\
    GPT3.5-Wiki      &     29,674 &     24,437 &     29,674 &     29,748 &    115,564 &    145,312 \\
    MS MARCO            &    389,304 &    389,974 &    389,987 &    416,763 &  3,006,996 &  3,422,436 \\
    Multilingual-NLI    &     64,900 &     99,809 &    100,752 &     68,096 &    743,857 &    811,884 \\
    \midrule
    Total & 969,988 & 823,490 & 1,022,299 & 1,037,163 & 7,039,826 & 8,074,916 \\
    \bottomrule
\end{tabular}
\caption{Basic statistics of the training datasets. \emph{\# Questions} refers to the number of unique questions with at least one positive/negative passage. \emph{MAUPQA} is a collection of seven individual datasets, but we provide aggregate counts for brevity. \emph{Total} represents the concatenation of all datasets.}
\label{tab:basic-stats}
\end{table*}

\section{Silver Retriever}

\subsection{Training Datasets}
\label{sec:dataset}
We use a collection of four existing and five newly created datasets for training. We summarize their basic statistics in Table \ref{tab:basic-stats} and give a short description of each of them below.

\subsubsection{Existing datasets}

\paragraph{PolQA} \citep{polqa} is the first Polish dataset for open-domain question answering. It consists of 7,000 questions from the trivia domain and 87,525 manually matched Wikipedia passages. We use the training split of the dataset.

\paragraph{MAUPQA} \citep{rybak-2023-maupqa} is a collection of seven datasets for Polish passage retrieval. Overall, it contains almost 400,000 question-passage pairs. The examples are weakly labeled, either automatically mined, generated, or machine-translated from English. The questions cover trivia, website FAQs, and various online texts.

\paragraph{PoQuAD} \citep{10.1145/3587259.3627548} is a Polish equivalent of the SQuAD \citep{rajpurkar-etal-2018-know}. It consists of more than 70,000 question-passage pairs. Contrary to the PolQA dataset, the dataset was created by asking annotators to write a question about a given Wikipedia passage. We use the training split of the dataset.

\paragraph{PolEval 2021 Pairs} During the 2021 PolEval Question Answering task \cite{ogrodniczukpoleval}, one of the task participants \cite{rybak2021retrieve} manually annotated over 10,000 question-passage pairs. 
We use it without any modification.

\subsubsection{New datasets}

\paragraph{1z10} \emph{Jeden z dziesięciu} (Eng. `One out of ten') is a popular Polish TV quiz show in which the contestants answer trivia questions about science, history, and the arts. We use the Whisper model \citep{radford2022whisper} to transcribe 333 episodes of the show and the GPT-3.5\footnote{\url{https://platform.openai.com/docs/models/gpt-3-5}} to extract question-answer pairs.

To match questions with relevant passages, we first use SpaCy \cite{spacy} to lemmatize questions and a corpus of Wikipedia passages distributed with the PolQA dataset. Then, the BM25 selects the top 100 passages which we re-rank using the mMiniLM-L6-v2 cross-encoder \cite{bonifacio2021mmarco}.\footnote{\url{https://hf.co/unicamp-dl/mMiniLM-L6-v2-mmarco-v2}} Finally, we select the top 5 passages and verify them with GPT-3.5. In total, we get 22,835 questions matched with 22,014 relevant passages. We keep the irrelevant passages as negatives.

\paragraph{GPT-3.5-CC and GPT-3.5-Wiki} Similar to the GenGPT3 dataset from MAUPQA, we prompt the generative language model to write a question about a given passage. However, we use GPT-3.5 instead of GPT-3 \citep{ouyangtraining}. We sample passages either from Polish Wikipedia or from the Polish part of CCNet \citep{wenzek-etal-2020-ccnet}. We generate nearly 30,000 questions for each source.

\paragraph{Polish MS MARCO} Similar to BEIR-PL \citep{wojtasik2023beirpl}, we translate the training subset of the MS MARCO dataset \citep{nguyen2016ms} into Polish. We use the in-house Allegro\footnote{\url{https://ml.allegro.tech}} machine translation model. We only consider relevant question-passage pairs and ignore existing hard negatives. Instead, we mine for the hard negatives ourselves (see Section \ref{sec:negatives}).

\paragraph{Multilingual NLI} We convert the collection of machine-translated natural language inference (NLI) datasets \citep{laurer_less_2022} into question-passage pairs. We convert the \emph{premise} to a question by adding the prefix `Czy' (Eng. `Does') and use the \emph{hypothesis} as a passage. Then, we consider \emph{entailment} and \emph{contradiction} labels as relevant (with answers `Yes' and `No' respectively) and \emph{neutral} as negative.

\subsection{Hard negatives}
\label{sec:negatives}
Hard negatives are negative (irrelevant) question-passage pairs that are difficult for the model to distinguish from the relevant pair, often due to the high lexical overlap between the question and the passage. Such pairs are crucial for training robust neural retrievers \citep{karpukhin-etal-2020-dense}.

We follow a similar strategy to \citet{ren-etal-2021-rocketqav2} for mining hard negatives. First, we lemmatize the questions and the corpus of passages and use the BM25 to select the top 10 candidate passages for each question. Then, we score them using the mMiniLM-L6-v2 cross-encoder and keep only the irrelevant passages.

We add hard negatives to all datasets except PolQA, which already contains hard negatives. As a source of passages, we use either the Wikipedia passages distributed with the PolQA dataset (for datasets that already use it) or the other passages of a given dataset (for all other datasets).

\subsection{Denoising}
\label{sec:denoising}
To increase the quality of the training pairs, we apply several steps of filtering noisy pairs.\footnote{We publish the script for filtering at: \url{https://github.com/360er0/silver-retriever}} We remove questions/passages that are too short or too long, passages that are relevant to too many questions, and passages that are too similar to their questions. We also score pairs using the E5-Base bi-encoder and mT5-3B cross-encoder\footnote{\url{https://hf.co/unicamp-dl/mt5-3B-mmarco-en-pt}} to remove relevant pairs with scores lower than 10\% and negative pairs with scores higher than 90\%. Finally, we manually create a blacklist for both entire questions and individual words. In total, we discard 14\% of the relevant question-passage pairs.

\subsection{Training}
\label{sec:model}
We use a standard dense passage retriever (DPR, \citealp{karpukhin-etal-2020-dense}) architecture implemented in the Tevatron library \citep{Gao2022TevatronAE}. We fine-tune HerBERT Base model \citep{mroczkowski-etal-2021-herbert} for 15,000 steps, with batch size 1,024 and learning rate $2 \cdot 10^{-5}$. We leave the rest of the hyperparameters at their default values. We use all training datasets (see Section \ref{sec:dataset}), but drop questions without relevant passages.

\begin{table*}[!ht]
\renewcommand*{\arraystretch}{1.05}
\setlength{\tabcolsep}{6.5pt}
\centering
\begin{tabular}{l|rr|rr|rr|rr}
    \toprule
    \multirow{2}{*}{\bf{Model}} & \multicolumn{2}{c|}{\bf{PolQA}} & \multicolumn{2}{c|}{\bf{Allegro FAQ}} & \multicolumn{2}{c|}{\bf{Legal Questions}} & \multicolumn{2}{c}{\bf{Average}} \\
    & \bf{Acc} & \bf{NDCG} & \bf{Acc} & \bf{NDCG} & \bf{Acc} & \bf{NDCG} & \bf{Acc} & \bf{NDCG}\\
    \midrule
    \multicolumn{9}{c}{\bf{Lexical models}} \\
    \midrule
    BM25 & 61.35 & 24.51 & 66.89 & 48.71 & \bf{96.38} & \bf{82.21} & 74.87 & 51.81 \\
    BM25 (lemma) & 71.49 & 31.97 & 75.33 & 55.70 & 94.57 & 78.65 & 80.46 & 55.44 \\
    \midrule
    \multicolumn{9}{c}{\bf{Multilingual models}} \\
    \midrule
    MiniLM-L12-v2 & 37.24 & 11.93 & 71.67 & 51.25 & 78.97 & 54.44 & 62.62 & 39.21 \\
    LaBSE & 46.23 & 15.53 & 67.11 & 46.71 & 81.34 & 56.16 & 64.89 & 39.47 \\
    mContriever-Base & 78.66 & 36.30 & 84.44 & 67.38 & 95.82 & 77.42 & 86.31 & 60.37 \\
    E5-Base & 86.61 & \bf{46.08} & 91.89 & 75.90 & 96.24 & 77.69 & 91.58 & 66.56 \\
    \midrule
    \multicolumn{9}{c}{\bf{Polish models}} \\
    \midrule
    ST-DistilRoBERTa & 48.43 & 16.73 & 84.89 & 64.39 & 88.02 & 63.76 & 73.78 & 48.29 \\
    ST-MPNet & 56.80 & 21.55 & 86.00 & 65.44 & 87.19 & 62.99 & 76.66 & 49.99 \\
    HerBERT-QA & 75.84 & 32.52 & 85.78 & 63.58 & 91.09 & 66.99 & 84.23 & 54.36 \\
    Silver Retriever (ours) & \bf{87.24} & 43.40 & \bf{94.56} & \bf{79.66} & 95.54 & 77.10 & \bf{92.45} & \bf{66.72} \\
    \bottomrule
\end{tabular}
\caption{Performance of passage retrieval for lexical and neural models.  We use top 10 accuracy and NDCG@10 on the test splits of each dataset and bold the highest score for each metric.} 
\label{tab:results}
\end{table*}

\section{Evaluation}
\label{sec:eval}
We evaluate the Silver Retriever on three passage retrieval datasets: PolQA, Allegro FAQ, and Legal Questions. The last two datasets were introduced in the 2022 PolEval Passage Retrieval task \citep{kobylinski2023poleval}. We compare the performance of our model with other state-of-the-art Polish and multilingual models.

For all tasks, we use both Accuracy@10 (i.e., there is at least one relevant passage within the top 10 retrieved passages) and NDCG@10 (i.e., the score of each relevant passage within the top 10 retrieved passages depends descendingly on its position, \citealp{ndcg}).

\subsection{PolQA}
We use the test subset of the PolQA dataset, and unlike to the evaluation in the original work, we use all passages annotated as relevant (not just those found by human annotators). This gives a better estimate of the true performance but may overestimate the results for the BM25 baselines.\footnote{In the original work, the passages were found either by human annotators or by the BM25. As a result, all BM25 predictions were manually verified which is not the case for other models. The retrieved passages could be relevant but were not labeled as such.} Similar, to the original evaluation, we ignore questions without relevant passages.

\subsection{Allegro FAQ}
Allegro FAQ\footnote{\url{https://hf.co/datasets/piotr-rybak/allegro-faq}} is a dataset of 900 frequently asked questions and 921 help articles about the large Polish e-commerce platform - Allegro.com. Each question-passage pair is manually checked and edited if necessary.

\subsection{Legal Questions}
Legal Questions\footnote{\url{https://hf.co/datasets/piotr-rybak/legal-questions}} consists of 718 questions and 26,000 passages extracted from over 1,000 acts of law. We ignore the title of the passage because it degrades performance on all tested models.

\subsection{Results}
\label{sec:results}

The BM25 retrievers, especially when using lemmas instead of word forms, achieve competitive performance on PolQA and outperform many neural models. However, the best model is either Silver Retriever (considering the accuracy of 87.24\%) or E5-Base (NDCG of 46.08\%).

For the Allegro FAQ, Silver Retriever scores much higher than E5-Base for both accuracy (94.56\% vs 91.89\%) and NDCG (79.66\% vs 76.02\%). The BM25 performs better than LaBSE, and MiniLM-L12-v2, but worse than the other models that achieve similar performance (around 85\% accuracy and 65\% NDCG).

For the Legal Questions, the BM25 achieves the best results (96.38\% accuracy and 82.21\% NDCG) and outperforms all neural models. This is not surprising since the passage length is often longer than 512 tokens and there is a high lexical overlap between questions and passages. Apart from the BM25 models, the best neural retriever is E5-Base, which is on par with BM25 in terms of accuracy (96.24\% vs 96.38\%) but achieves a much lower NDCG (77.79\% vs 82.21\%).

On average, the SilverRetriver performs best in both accuracy (92.45\%) and NDCG (66.72\%). The E5-Base model is a close second (91.63\% and 66.63\% respectively). The other models perform much worse, often worse than the BM25.

\section{Ablation Study}
\label{sec:ablation}
We analyze the impact of training choices using only the PolQA validation set since two other datasets only contain a test set. Compared to the main training (see Section \ref{sec:denoising}), we fine-tuned the HerBERT Base model for 5,000 steps, and a batch size of 256. The effect of adding hard negatives is positive but minimal (probably due to the short training). Denoising improves both accuracy and NDCG by 1.6 p.p. The effect of a larger batch size (1024 vs. 256) is also positive for both metrics.

\begin{table}[!ht]
\renewcommand*{\arraystretch}{1.05}
\setlength{\tabcolsep}{6pt}
\centering
\begin{tabular}{l|rr}
    \toprule
    \bf{Variant} & \bf{Acc@10} & \bf{NDCG@10} \\
    \midrule
    base & 79.09 & 35.51 \\
    + hard negatives & 79.20 & 36.64 \\
    + denoising & 80.89 & 38.28 \\
    + batch size of 1024 & \bf{81.31} & \bf{39.13} \\
    \bottomrule
\end{tabular}
\caption{Impact of different design choices on the performance on the PolQA validation dataset. Subsequent models include all previous changes.}
\label{tab:ablation}
\end{table}

\section{Conclusion}
\label{sec:conclusion}

We present Silver Retriever, a neural passage retriever designed for Polish and trained on a variety of manually and weakly labeled datasets. The model's performance is evaluated on three retrieval tasks, on which it achieves strong results compared to several Polish and multilingual models, with the best scores on the Allegro FAQ and the best average score across all tasks.

In addition, the article contributes to the research community by open-sourcing five new passage retrieval datasets with a total of 500,000 questions.

In summary, the Silver Retriever and associated datasets represent a significant advance in the field of passage retrieval for Polish, and provide valuable resources for improving the accuracy and efficiency of open-domain question answering systems.

\section{Acknowledgments}
We thank the Allegro.com Machine Learning Research Team for giving us access to their machine translation model and Konrad Kaczyński for helping us transcribe the \emph{Jeden z dziesięciu} episodes using the Whisper model.

This work was supported by the European Regional Development Fund as a part of 2014–2020 Smart Growth Operational Programme, CLARIN — Common Language Resources and Technology Infrastructure, project no. POIR.04.02.00-00C002/19.

\nocite{*}
\section{Bibliographical References}\label{reference}
\bibliographystyle{lrec_natbib}
\bibliography{lrec-coling2024-example}


\end{document}